# Semi-analytical approximations to statistical moments of sigmoid and softmax mappings of normal variables

*J. Daunizeau*[1,2]

[1] Brain and Spine Institute, Paris, France

[2] ETH, Zurich, France

Address for correspondence:

Jean Daunizeau

Motivation, Brain and Behaviour Group

Brain and Spine Institute (ICM), INSERM UMR S975.

47, bvd de l'Hopital, 75013, Paris, France.

Tel: +33 1 57 27 43 26

Fax: +33 1 57 27 47 94

Mail: jean.daunizeau@gmail.com

Web: https://sites.google.com/site/jeandaunizeauswebsite/

This note is concerned with accurate and computationally efficient approximations of moments of Gaussian random variables passed through sigmoid or softmax mappings. By definition, a sigmoid function is real-valued and differentiable, having a non-negative first derivative which is bell shaped (Han and Moraga, 1995). They owe their popularity to the fact that they capture a canonical form of non-linearity, namely: steps. Applications of sigmoid mappings include, but are not limited to, the description of learning dynamics (see, e.g., Leibowitz et al., 2010) and psychometric functions (see, e.g., Wichmann and Hill, 2001), input/output relationships of artificial (see, e.g., Jain et al., 1996) and natural neurons (see, e.g., Marreiros et al., 2008), cumulative distribution functions of usual probability density functions (see, e.g., Waissi and Rossin, 1996), logistic regression (see, e.g., Hilbe, 2009) and machine learning (see, e.g., Witten et al., 2016), etc. Although very simple mathematical objects, they challenge standard analytical approximation schemes such as Taylor and/or Fourier series. In particular, the fact that their Taylor series converge very slowly is problematic for deriving approximations of moments of Gaussian variables passed through sigmoid mappings. This is the central issue of this note, which we address using a semi-analytical approach.

We will use Monte-Carlo sampling to evaluate the quality of the proposed approximations, and compare them to standard second-order Taylor approximations. *En passant*, we derive analytical properties of sigmoid and softmax mappings (as well as their log-transform), in terms of their gradients and Hessians. We also expose a few niche applications of these approximations, which mainly arise in the context of variational Bayesian inference (Beal, 2003; Daunizeau et al., 2009; Friston et al., 2007; Smídl and Quinn, 2006) on specific generative models.

# 1. The sigmoid mapping

In what follows, $x$ is a random variable that is normally distributed, with mean $\mu$ and variance $\Sigma$, i.e.: $x \sim N(\mu, \Sigma)$, where $N(\cdot, \cdot)$ denotes the gaussian probability density function.

## 1.0. On the logistic distribution

In the following, we will rely heavily on a moment-matching approach from the normal to the logistic probability density functions (and back). Let us recall that the cumulative density function of the logistic distribution is simply a sigmoid mapping (Balakrishnan, 2013):

$$P(z \leq x) = \frac{1}{1 + \exp\left(-\frac{x-t}{\rho}\right)} \Rightarrow \begin{cases} E[z] = t \\ V[z] = \frac{\pi^2}{3} \rho^2 \end{cases}$$

This can be directly used as an analytical approximation to the cumulative density function of the normal density with mean $t$ and variance $\frac{\pi^2}{3} \rho^2$.

## 1.1. The expected sigmoid mapping

Let us first consider the canonical sigmoid mapping:

$$s : x \to s(x) = \frac{1}{1 + \exp(-x)} \tag{1}$$

Its derivatives are given by:

$$s'(x) = \frac{\exp(-x)}{(1+\exp(-x))^2}$$
$$= s(x)(1-s(x))$$
$$s''(x) = \frac{\partial}{\partial x}\left[s(x)(1-s(x))\right] = s'(x) - 2s'(x)s(x) = s'(x)(1-2s(x)) \quad (2)$$
$$= s(x)(1-s(x))(1-2s(x))$$
$$\frac{\partial^n}{\partial x^n} s(x) = s(x)\prod_{i=1}^{n}(1-is(x))$$

Equation 2 can be used to derive a simple approximation to the expected sigmoid mapping $E[s(x)] \triangleq \langle s(x) \rangle$, based upon a second-order Taylor expansion:

$$\langle s(x) \rangle = \left\langle s(\mu) + s'(\mu)(x-\mu) + \frac{1}{2}s''(\mu)(x-\mu)^2 + ... \right\rangle$$
$$\approx s(\mu)\left[1 + \frac{1}{2}(1-s(\mu))(1-2s(\mu))\Sigma\right] \quad (3)$$

Under this approximation, $\langle s(x) \rangle$ is always positive, but it can be greater than one, which can be problematic when using the sigmoid mapping as a probabilistic statement. In the latter case, one may want to truncate the Taylor expansion to first order, or use another approximation.

Note that such an approximation should conform to intuition, namely that (i) the expected sigmoid should be positive and smaller than one, (ii) it should be a sigmoidal function of $\mu$, (iii) its slope should be a decreasing function of $\Sigma$, (iv) at the high precision limit $\mu/\sqrt{\Sigma} \to \pm\infty$, it should converge to $s(\mu)$.

We thus propose the following fixed-form approximation:

$$\langle s(x) \rangle \approx s\left(\frac{\mu}{\sqrt{1+a\Sigma}}\right) \quad (4)$$

which conforms with the four above desiderata. Figure 1 above depicts the comparison between this approximation (with fitted parameter $\hat{a} \approx 0.368$) and Monte-Carlo estimates of $\langle s(x) \rangle$ over a wide range of first- and second-order moments of the distribution of $x$ (i.e.: $\mu \in [-10, 10]$ and $\Sigma \in [2^{-4}, 2^{8}]$).

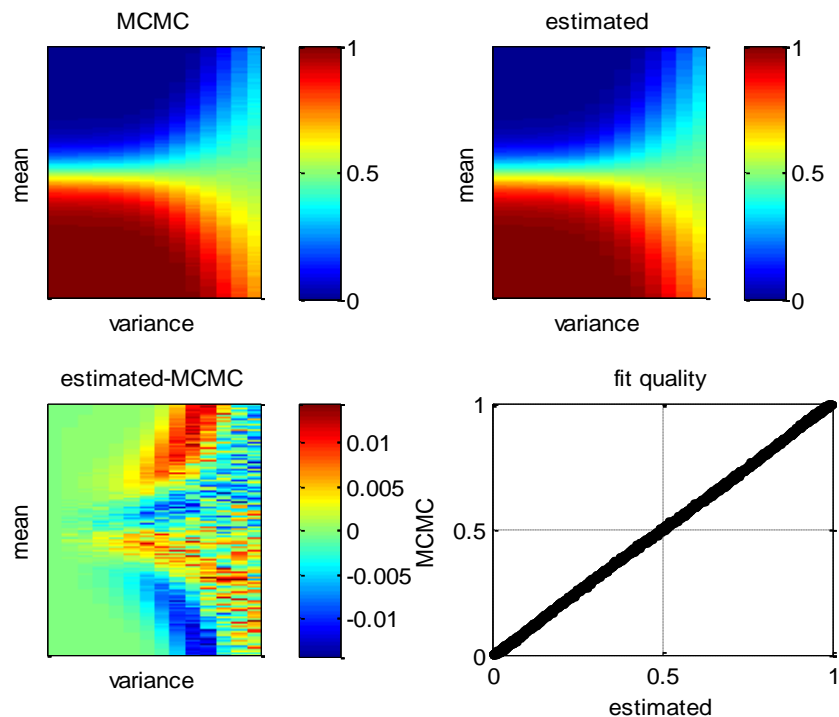

Figure 1. Fixed-form approximation to the expected sigmoid

Even though some structured error is visible, one can see that the relative approximation error is smaller than 2%.

For comparison purposes, Figure 2 below depicts the first- and second-order Taylor approximations of the expected sigmoid, over the same range of moments of the distribution of $x$.

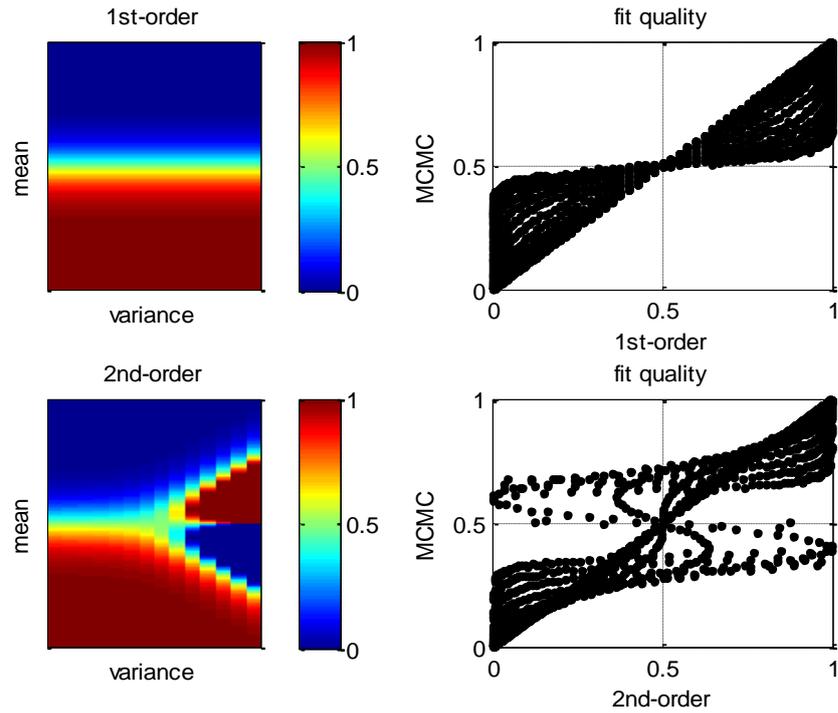

Figure 2. Taylor approximations to the expected sigmoid

One can immediately see how the respective quality of the fixed-form and the Taylor approximations compare to each other. In brief, the first-order Taylor approximation shows (obviously) no effect of $\Sigma$, and the second order Taylor approximation strongly departs from a monotonic sigmoidal behaviour when $\Sigma$ increases. In turn, both first- and second-order Taylor approximation yields much higher approximation errors. This serves as a numerical motivation for using the above fixed-form approximation, for which we will now try to give an intuitive derivation.

First, let us recall that the (scaled) sigmoid mapping is related to the logistic cumulative density function, i.e.:

$$\frac{1}{1+\exp\left(-\dfrac{x}{\rho}\right)} = P(z \geq 0 | x) \quad \text{with:} \quad \begin{cases} E[z|x] = x \\ V[z|x] = \dfrac{\pi^2}{3}\rho^2 \end{cases} \tag{5}$$

where $p(z|x)$ is a logistic probability density function with first- and second-order moments as above (i.e., $x$ is the first-order moment of the random variable $z$). Let $q(x)$ be the (Gaussian) probability density function of $x$, with first- and second-order moments $\mu$ and $\Sigma$, respectively. The expression for the expected sigmoid under $q$ is given by:

$$\begin{aligned}
\langle s(x) \rangle &= \int_{-\infty}^{\infty} P(z \geq 0 | x) q(x) dx \\
&= \int_{-\infty}^{\infty} \left[ \int_0^{\infty} p(z|x) dz \right] q(x) dx \\
&= \int_0^{\infty} \left[ \int_{-\infty}^{\infty} p(z|x) q(x) dx \right] dz \\
&= \int_0^{\infty} p(z) dz \\
&= P(z \geq 0)
\end{aligned} \tag{6}$$

where no approximation has been necessary so far.

Now, matching the first- and second-order moments of the logistic (conditional) density of $z|x$, yields the following Gaussian approximate marginal density of $z$:

$$\left. \begin{aligned} z|x &\sim N\left(x, \frac{\pi^2}{3}\right) \\ x &\sim N(\mu, \Sigma) \end{aligned} \right\} \Rightarrow z \sim N\left(\mu, \frac{\pi^2}{3} + \Sigma\right) \tag{7}$$

Reverting the moment matching Gaussian approximation back to the logistic density then yields:

$$P(z \geq 0) = \frac{1}{1+\exp\left(-\frac{\mu}{\sqrt{1+\frac{3}{\pi^2}\Sigma}}\right)} = s\left(\frac{\mu}{\sqrt{1+\frac{3}{\pi^2}\Sigma}}\right) \approx \langle s(x) \rangle \qquad (8)$$

which has the desired functional form. Note that $3/\pi^2 \approx 0.304$, which is close (but not exactly equal) to the best fit parameter $\hat{a}$ above. This is because the double moment matching approximation is slightly overconfident, which has to be compensated for by a slight rescaling of the variance of $x$.

## 1.2. The expected log-sigmoid mapping

Now let us look at the properties of the log-sigmoid mapping:

$$\log s(x) = x - \log(1+\exp(x))$$

$$\Rightarrow \begin{cases} \log(1-s(x)) = \log\left(\frac{1}{1+\exp(x)}\right) = -\log(1+\exp(x)) \\ \qquad\qquad\qquad = \log s(x) - x \\ \frac{\partial}{\partial x}[\log s(x)] = 1 - \frac{\exp(x)}{1+\exp(x)} \\ \qquad\qquad\qquad = 1 - s(x) \\ \frac{\partial^2}{\partial x^2}[\log s(x)] = s(x)(s(x)-1) \end{cases} \qquad (9)$$

NB: one can see from equation 9 that the log-sigmoid mapping is related to the antiderivative of the sigmoid mapping.

Using a second-order Taylor expansion, equation 9 directly yields an approximate expectation of the log-sigmoid mapping:

$$\langle \log s(x) \rangle \approx \log s(\mu) - \frac{1}{2} s(\mu)(1 - s(\mu)) \Sigma \qquad (10)$$

Note that the above approximation does not ensure a proper normalization, i.e.:

$\exp\langle \log s(x) \rangle + \exp\langle \log(1 - s(x)) \rangle = 1$ may not be satisfied...

However, imposing an ad-hoc normalization constraint upon the resulting sigmoidal mapping may not be desirable, since this unfortunately yields the first-order Taylor approximation:

$$\begin{aligned} 1 &= \exp\langle \log s(x) \rangle + \exp\langle \log(1 - s(x)) \rangle \\ &= \exp\langle \log s(x) \rangle (1 + \exp\langle -x \rangle) \\ \Leftrightarrow \langle \log s(x) \rangle &= \log \frac{1}{1 + \exp\langle -x \rangle} \end{aligned} \qquad (11)$$

We thus know that the exact expected log-sigmoid mapping does not normalize. Having said this, equation 10 is still not satisfactory because applying the exponential map to it does not satisfy the intuitive properties of a sigmoidal mapping (e.g., $\exp\langle \log s(x) \rangle$ can be greater than one).

Thus, we propose to use the following fixed-form approximation:

$$\langle \log s(x) \rangle \approx \log s\left( \frac{\mu + b\Sigma^c}{\sqrt{1 + a\Sigma^d}} \right) \qquad (12)$$

where we fit the parameters ($\hat{a} \approx 0.205$, $\hat{b} \approx -0.319$, $\hat{c} \approx 0.781$ and $\hat{d} \approx 0.870$) to Monte-Carlo estimates of the expected log-sigmoid, over the same range of first- and second-order moments of the distribution of $x$. Note that the fixed-form approximation in equation 12 satisfies the above desiderata (i), (ii), (iii) and (iv) on the sigmoid mapping.

Figure 3 below depicts the comparison of the above fixed form approximation to the Monte-Carlo estimate:

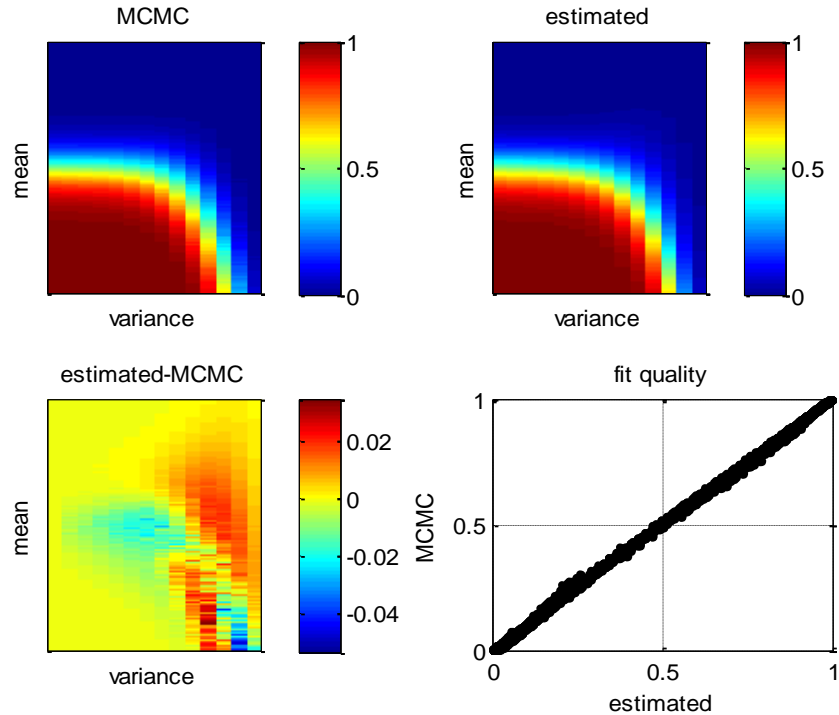

Figure 3: Fixed-form approximation to the expected log-sigmoid

Note that we have chosen to display the effective sigmoidal mappings $\exp\langle\log s(x)\rangle$, so that the comparison to figures 1 and 2 is directly possible. In particular, one can see how the variance biases the effective sigmoid mapping $\exp\langle\log s(x)\rangle$. More precisely, increasing the variance $\Sigma$ eventually shifts the effective sigmoid's inflexion point upwards, i.e. the value of $\mu$ such that the effective sigmoid reaches $1/2$ increases with $\Sigma$. Note that this does not arise for $\langle s(x)\rangle$ (cf. Figure 1), where $\Sigma$ only rescales the sigmoid's slope (which also happens here, although to a lesser extent).

For comparison purposes, Figure 4 below depicts the expected log-sigmoid mapping, when derived using first- and second-order Taylor approximations:

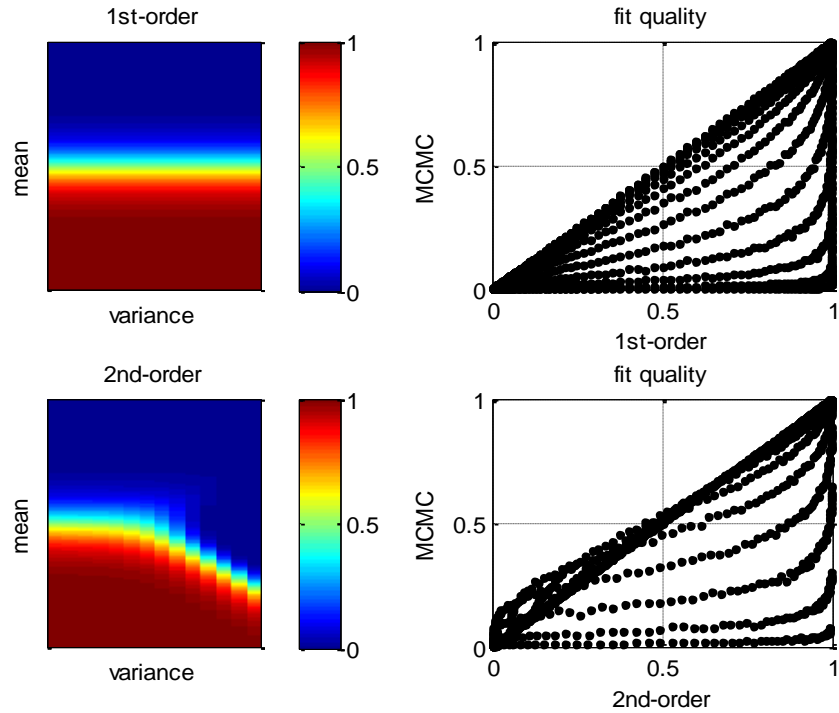

Figure 4: Taylor approximations to the expected log-sigmoid

One can see that the relative error is much higher than for the fixed-form approximation, and reaches about 100% for high variances $\Sigma$. Again, we take this as a numerical validation of the proposed fixed-form approximation, which yields a much lower relative approximation error (about 5%).

## 1.3. Extension to parametric sigmoidal mappings

Do these approximations extend to modified forms of the sigmoid mapping?

Let us recall the standard parametric form $s_0$ of the sigmoid mapping:

$$s_0(x) = \frac{1}{1+\exp\left(-\frac{x-t}{\rho}\right)} \tag{13}$$

where $t$ is the sigmoid's inflexion point (it is such that $s_0(t) = 1/2$) and $\rho$ is the sigmoid's slope (it controls how steep is the sigmoid).

It is trivial to see that $s_0$ is nothing more than the composition of the affine mapping $x \to (x-t)/\rho$ with the simple sigmoid $s$, i.e.: $s_0(x) = s((x-t)/\rho)$. This means that:

- Derivatives of any order obtain from applying the chain rule:

$$\left.\frac{\partial^n s_0}{\partial x^n}\right|_x = \frac{1}{\rho^n} \left.\frac{\partial^n s}{\partial x^n}\right|_{\frac{x-t}{\rho}}$$

$$= \frac{1}{\rho^n} s\left(\frac{x-t}{\rho}\right) \prod_{i=1}^{n}\left(1 - is\left(\frac{x-t}{\rho}\right)\right)$$

(14)

- Expectations of such sigmoid mappings, using the moment-matching-and-back trick above, obtain from first applying an affine transformation of variables ($\mu \to (\mu-t)/\rho$ and $\Sigma \to \Sigma/\rho^2$):

$$\langle s_0(x) \rangle = \left\langle s\left(\frac{x-t}{\rho}\right) \right\rangle$$

$$\approx s\left(\frac{\mu - t}{\sqrt{\rho^2 + \frac{3}{\pi^2}\Sigma}}\right)$$

(15)

and similarly for the expected log-sigmoid.

Another extension to the sigmoid mapping, which arises naturally in the context of (variational) Bayesian inference with log-normal priors on precision hyperparameters is the mapping: $x \to 1/(a + \exp(-x))$. It turns out that this form is but a particular case of the form $s_0$ above: $x \to s(x + \log a)/a$, for which we can obtain derivatives (equation 14) and accurate expectations (equation 15).

## 1.4. The variance of the sigmoid mapping

Let us now derive an analytical approximation to the variance $V_s$ of $s(x)$:

$$V_s = \left\langle \left( s(x) - \left\langle s(x) \right\rangle \right)^2 \right\rangle \tag{16}$$

First, let note that:

$$\begin{aligned} V_s &= \left\langle s(x)^2 \right\rangle - \left\langle s(x) \right\rangle^2 \\ &= \left\langle s(x) - s(x)(1 - s(x)) \right\rangle - \left\langle s(x) \right\rangle^2 \\ &= \left\langle s(x) \right\rangle - \left\langle s'(x) \right\rangle - \left\langle s(x) \right\rangle^2 \\ &= \left\langle s(x) \right\rangle \left( 1 - \left\langle s(x) \right\rangle \right) - \left\langle s'(x) \right\rangle \end{aligned} \tag{17}$$

The only term that is new here is $\left\langle s'(x) \right\rangle$. Some intuition may be derived from the logistic probability density function by recalling that $s'(x)$ can be understood as an almost Gaussian probability density function with the following mean and variance:

$$x \sim s'(x) \Rightarrow \begin{cases} E[x] = 0 \\ V[x] = \dfrac{\pi^2}{3} \end{cases} \tag{18}$$

Applying the moment-matching trick thus yields:

$$\begin{aligned} \left\langle s'(x) \right\rangle &= \int_{-\infty}^{\infty} N\left( 0, \frac{\pi^2}{3} \right) q(x) dx \\ &= \frac{\sqrt{3}}{2\pi^2 \sqrt{\Sigma}} \int_{-\infty}^{\infty} \exp f(x) dx \\ f(x) &= -\frac{3}{2\pi^2} x^2 - \frac{1}{2\Sigma} (x - \mu)^2 \end{aligned} \tag{19}$$

In Equation 19, $f(x)$ is a quadratic polynomial that can thus be written as:

$$f(x) = f(x_0) + \frac{1}{2} f''(x_0)(x-x_0)^2$$

$$x_0 = \frac{\mu}{\frac{3\Sigma}{\pi^2}+1}$$

$$f(x_0) = -\frac{1}{2} \frac{\mu^2}{\Sigma + \frac{\pi^2}{3}}$$

$$f''(x_0) = -\frac{3}{\pi^2} - \frac{1}{\Sigma}$$

(20)

where $x_0$ is the extremum of $f(x)$. Inserting Equation 20 into Equation 19 yields:

$$\langle s'(x) \rangle = \frac{\sqrt{3}}{2\pi^2 \sqrt{\Sigma}} \exp f(x_0) \int_{-\infty}^{\infty} \exp \frac{1}{2} \underbrace{f''(x_0)}_{<0}(x-x_0)^2 \, dx$$

$$= \frac{\sqrt{3}}{2\pi^2 \sqrt{\Sigma}} \exp f(x_0) \sqrt{\frac{2\pi}{-f''(x_0)}}$$

$$= \frac{1}{\sqrt{2\pi}} \frac{1}{\sqrt{\Sigma + \frac{\pi^2}{3}}} \exp\left(-\frac{1}{2} \frac{\mu^2}{\Sigma + \frac{\pi^2}{3}}\right)$$

(21)

where the second line derives from noting that the integrand has the form of an unnormalized Gaussian density. At this point, we can revert the moment-matching trick back by noting that Equation 21 has the form of a Gaussian probability density function with mean $\mu$ and variance $\Sigma + \pi^2/3$ evaluated at 0, i.e.:

$$\langle s'(x) \rangle = s'\left(\frac{\mu}{\sqrt{1 + \frac{3}{\pi^2}\Sigma}}\right)$$

(22)

Inserting Equation 22 into Equation 17 now yields the final analytical approximation to the variance of the sigmoid mapping:

$$V_s(\mu,\Sigma) \approx s\left(\frac{\mu}{\sqrt{1+\frac{3}{\pi^2}\Sigma}}\right)\left(1-s\left(\frac{\mu}{\sqrt{1+\frac{3}{\pi^2}\Sigma}}\right)\right)\left(1-\frac{1}{\sqrt{1+\frac{3}{\pi^2}\Sigma}}\right) \tag{23}$$

Note that the variance of the sigmoid is a monotically increasing function of $\Sigma$, and it is bounded between 0 (at the limit $\Sigma \to 0$) and $1/4$ (at the limit $\Sigma \to \infty$), i.e.:

$$\left.\begin{array}{l}\frac{\partial}{\partial \Sigma}V_s \geq 0 \\ 0 \leq V_s \leq \frac{1}{4}\end{array}\right\}\forall \mu, \forall \Sigma \quad , \begin{cases} V_s \xrightarrow{\Sigma \to 0} 0 \\ V_s \xrightarrow{\Sigma \to \infty} \frac{1}{4}\end{cases} \tag{24}$$

The quality of the analytical approximation can be eyeballed on Figure 5 below.

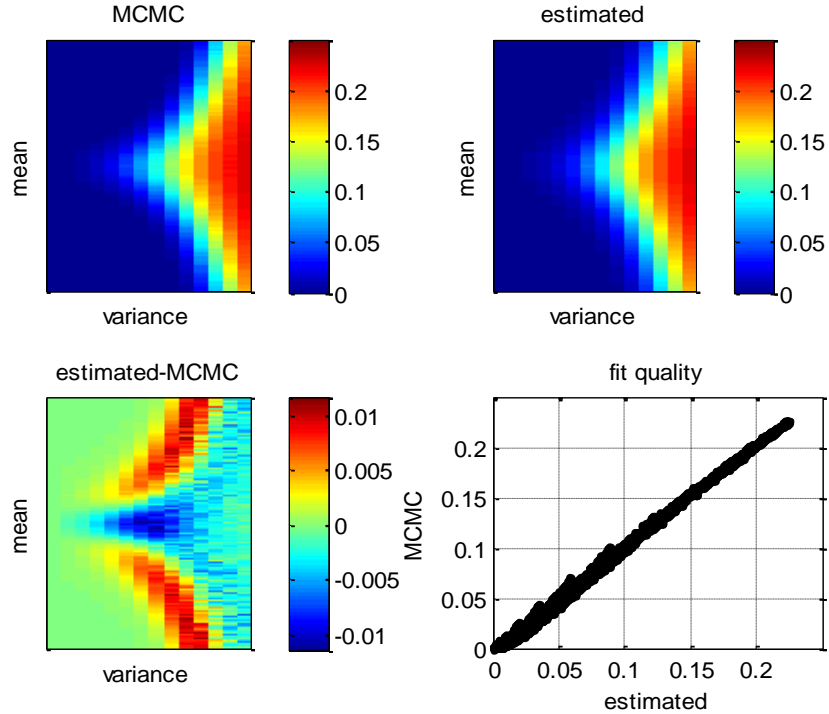

Figure 5: Fixed-form approximation to the variance of the sigmoid mapping

One can see that the variance of the sigmoid increases with $\Sigma$ and decreases when $\mu \to \pm\infty$. This is because the gradient of the sigmoid mapping tends to zero when $\mu \to \pm\infty$.

## 2. The softmax mapping

In what follows, $x$ is a vector-valued random variable that is normally distributed, with mean $\mu$ and variance-covariance matrix $\Sigma$, i.e.: $x \sim N(\mu, \Sigma)$, where $N(\cdot, \cdot)$ denotes the multivariate gaussian probability density function.

Let us now extend the above results to degenerate sigmoid mappings, e.g. *softmax* mappings. Let $\pi_k : x \times k \rightarrow \pi_k(x)$ be the standard softmax mapping, where $x$ is a vector of real values indexed by $k$:

$$\pi_k(x) \triangleq \frac{\exp(x_k)}{\sum_{k'} \exp(x_{k'})} \tag{25}$$

Note that softmax maps normalize, i.e. $1 = \sum_k \pi_k(x)$, irrespective of $x$.

Deriving the first- and second-order behaviour of the softmax mapping in a compact (matrix) form turns out to be much easier if one first derives the gradient and Hessian of the log-softmax. This is because:

$$\begin{aligned}
\frac{\partial \pi_k}{\partial x} &= \pi_k \frac{\partial}{\partial x}[\log \pi_k] \\
\frac{\partial^2 \pi_k}{\partial x^2} &= \pi_k \left( \frac{\partial^2}{\partial x^2}[\log \pi_k] + \frac{\partial}{\partial x}[\log \pi_k] \frac{\partial}{\partial x}[\log \pi_k]^T \right)
\end{aligned} \tag{26}$$

where equation 26 follows from having reversed the chain rule.

Thus, let us inspect the gradient of the log-softmax map:

$$\log \pi_k(x) = x_k - \log \sum_{k'} \exp(x_{k'})$$

$$\Rightarrow \begin{cases} \frac{\partial}{\partial x_k}[\log \pi_k(x)] = 1 - \frac{\exp(x_k)}{\sum_{k'} \exp(x_{k'})} = 1 - \pi_k(x) \\ \frac{\partial}{\partial x_{k'}}[\log \pi_k(x)] = -\frac{\exp(x_{k'})}{\sum_{k'} \exp(x_{k'})} = -\pi_{k'}(x) \end{cases} \Rightarrow \frac{\partial}{\partial x}[\log \pi_k(x)] = e_k - \pi(x) \tag{27}$$

where $e_k$ is the $k$th column of the identity matrix and $\pi(x)$ is a vector, whose entries are the softmax functions $\pi_k(x)$.

Similarly, its Hessian can be obtained as follows:

$$\frac{\partial^2}{\partial x^2}\left[\log \pi_k(x)\right] = -\frac{\partial}{\partial x}\pi(x)^T$$

$$= -\begin{bmatrix} \frac{\partial \pi_1}{\partial x_1} & \frac{\partial \pi_2}{\partial x_1} & \cdots \\ \frac{\partial \pi_1}{\partial x_2} & \frac{\partial \pi_2}{\partial x_2} & \\ \vdots & & \ddots \end{bmatrix} \quad (28)$$

$$= -\begin{bmatrix} \pi_1(1-\pi_1) & -\pi_1\pi_2 & \cdots \\ -\pi_1\pi_2 & \pi_2(1-\pi_2) & \\ \vdots & & \ddots \end{bmatrix}$$

$$= \pi(x)\pi(x)^T - Diag(\pi(x))$$

Interestingly, equation 28 implies that the Hessian of the log-softmax mapping $\log \pi_k(x)$ does not depend upon $k$. Note that equation 28 can be used to derive approximations of the expected log-softmax, based upon a second-order Taylor expansion:

$$\langle \log \pi_k(x) \rangle \approx \log \pi_k(E[x]) + \frac{1}{2}tr\left[\frac{\partial^2}{\partial x^2}\left[\log \pi_k(x)\right]\bigg|_{E[x]} V[x]\right]$$

$$= \log \pi_k(\mu) + \frac{1}{2}tr\left[\left(\pi(\mu)\pi(\mu)^T - Diag(\pi(\mu))\right)\Sigma\right] \quad (29)$$

which is really a direct generalization of the above equivalent relation for the sigmoid mapping. As before, the above approximation does not ensure a proper normalization, i.e. $1 = \sum_k \exp\langle \log \pi_k(x) \rangle$ may not be satisfied...

Anyway, the above expressions for the gradient and Hessian of the log-softmax yield those of the softmax mapping itself:

$$\frac{\partial \pi_k}{\partial x} = \pi_k(x)(e_k - \pi(x))$$

$$\frac{\partial^2 \pi_k}{\partial x^2} = \pi_k(x)\left(\pi(x)\pi(x)^T - Diag(\pi(x)) + (e_k - \pi(x))(e_k - \pi(x))^T\right)$$

(30)

which can be used again to obtain approximations for the expected softmax...

Note that, in the 2D case, the softmax is equivalent to the sigmoid mapping:

$$\pi_1(x) = \frac{\exp(x_1)}{\exp(x_1) + \exp(x_2)} = \frac{1}{1 + \exp(-(x_1 - x_2))} \Rightarrow \pi_1(x) = s(x_1 - x_2)$$

(31)

This means that one can understand the softmax mapping as the probability $P(z_1 > z_2 | x)$, where $z_1$ and $z_2$ are logistic random variates with means $x_1$ and $x_2$, respectively. One could thus use the same 'moment-matching-and-back' trick (see Equations 5-8 above), noting that:

$$\left.\begin{matrix} E[x] = \mu \\ V[x] = \Sigma \end{matrix}\right\} \Rightarrow \begin{cases} E[x_1 - x_2] = c\mu \\ V[x_1 - x_2] = c\Sigma c^T \end{cases} \text{ with } c = \begin{bmatrix} 1 & -1 \end{bmatrix}$$

(32)

where we refer to $c$ as the contrast vector. Taken together, equations 31 and 32 provide the following fixed-form approximation:

$$\langle \pi_1(x) \rangle = \langle s(cx) \rangle \approx s\left(\frac{c\mu}{\sqrt{1 + \frac{3}{\pi^2} c\Sigma c^T}}\right)$$

(33)

which will behave exactly as the above fixed-form sigmoid approximation. Unfortunately, this does not generalize to any arbitrary dimension. First, let us note that the softmax mapping can be expressed as a function of sigmoid mappings:

$$\pi_k(x) = \frac{1}{1 + \sum_{k' \neq k} \exp(-\tilde{x}_{k'})} = \frac{1}{2 - K + \sum_{k' \neq k} \frac{1}{s(\tilde{x}_{k'})}}$$

(34)

where we have used the compact notation: $\tilde{x}_{k'} \equiv x_k - x_{k'} = c_{k,k'} \cdot x$, with $c_{k,k'}$ being the associated contrast vector. Equation 34 reduces to equation 31 in the 2D-case. However, it can be seen that no simple interpretation of the form $P(z_k > z_1 \cap z_k > z_2 \cap ... \cap z_k > z_K | x)$ can be derived from Equation 34 in the general case. Nevertheless, we rely on Equation 34 to propose the following fixed-form approximation of the sigmoid mapping:

$$\langle \pi_k(x) \rangle \approx \frac{1}{2 - K + \sum_{k' \neq k} \frac{1}{\langle s(\tilde{x}_{k'}) \rangle}} \tag{35}$$

which is strictly true only in a two-dimensional problem (c.f. equations 31-33).

To evaluate the different approximations to the softmax mapping, we have conducted the following simulations. Let $x \sim N(\mu, \Sigma)$ be a three-dimensional vector that follows a normal density with mean $\mu$ and covariance matrix $\Sigma$ such that:

$$\mu_1 = 0 \text{ and } \Sigma = \sigma A A^T \text{ where } A = \begin{bmatrix} 1 & 0 & 0 \\ 0 & 1 & 0 \\ 0 & 0 & 1 \end{bmatrix} + \rho \begin{bmatrix} 0 & 1 & 1 \\ 1 & 0 & 1 \\ 1 & 1 & 0 \end{bmatrix}.$$

This construction of the covariance matrix $\Sigma$ ensures positive-definiteness and control over both the marginal variance $\sigma$ and the degree of correlation $d(\rho)$ between entries of the vector $x$.

Note that: (i) the contrasts $\tilde{x}_{k'}$ behave as Gaussian variates with variance $2\sigma(1 - d(\rho))$, and (ii) since (without loss of generality) we will be inspecting $\pi_1(x)$, we do not have to consider the correlation between $x_2$ and $x_3$.

Figure 6 below displays the first- and second- order Taylor approximations to the expected softmax mapping (to be compared with the fixed-form approximation, shown on Figure 7). This was done using $-\frac{1}{2}<\rho<1$ to span a marginal correlation $d(\rho)$ between -1 and 1, $0<\sigma<100$, and $-5<\{\mu_2 \text{ or } \mu_3\}<5$, with $\mu_1=0$ (which does not matter).

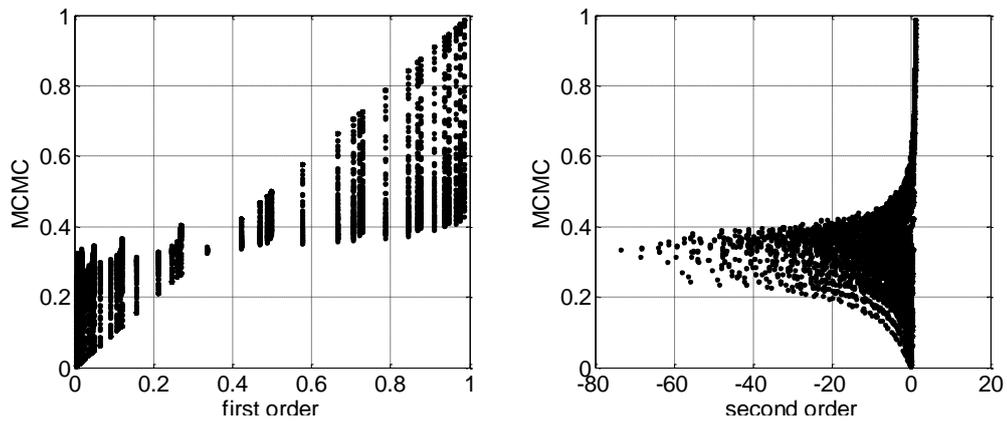

Figure 6. First- and second- order Taylor approximation to $\langle \pi_1(x) \rangle$

One can see that the relative error is about 50% and 10000% for the first- and second-order Taylor approximations, respectively. In contradistinction, the relative error remains below 2% for the fixed-form approximation, which is depicted on Figure 7 below:

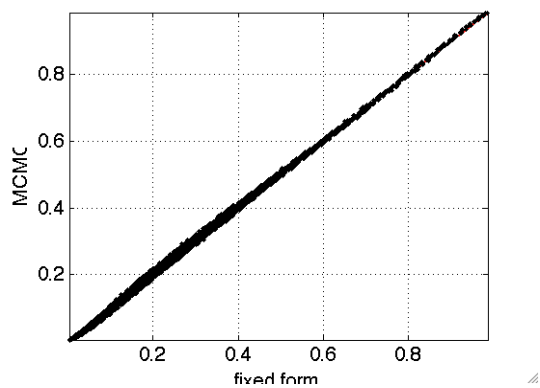

Figure 7. Fixed-form approximation to $\langle \pi_1(x) \rangle$

## 3. A few niche applications

As we will see below, analytical approximations on moments of sigmoids and log-sigmoids have direct though potentially unforeseen applications.

**3.1 Cumulative distribution function of the skewed normal distribution**

Skewed normal distributions arise in, e.g., stochastic processes that are bounded by an absorbing barrier or threshold (Anděl et al., 1984). An example of such processes is provided by drift-diffusion models of choice and reaction time data in the context of decision making (Ratcliff et al., 1999).

Let $N_s(x)$ be the normalized product between a sigmoid mapping and a gaussian probability density function:

$$N_s(x) = \frac{1}{K} s\left(\frac{x-t}{\rho}\right) N(x|\mu,\sigma)$$

$$K = \int_{-\infty}^{\infty} s\left(\frac{x-t}{\rho}\right) N(x|\mu,\sigma) dx \qquad (36)$$

Formally speaking, the probability density function of the skewed normal distribution can be obtained at the limit $\rho \to 0$. Thus, by abuse of notation, we will refer to $N_s(x)$ as the skewed normal probability density function. Note that this formulation was first introduced in the context of Bayesian estimation subject to uncertainty about parameter constraints (O'Hagan and Leonard, 1976).

Let $P_s(z)$ be the cumulative density function of the skewed normal distribution:

$$P_s(z) \triangleq \int_{-\infty}^{z} N_s(x) dx$$

$$= \frac{1}{K} \int_{-\infty}^{z} s\left(\frac{x-t}{\rho}\right) N(x|\mu,\sigma) dx$$

$$= \frac{1}{K}\left[ s\left(\frac{x-t}{\rho}\right) \int_{-\infty}^{x} N(x'|\mu,\sigma) dx' \right]_{-\infty}^{z} - \frac{1}{K} \int_{-\infty}^{z} s'\left(\frac{x-t}{\rho}\right) N(x|\mu,\sigma) dx \qquad (37)$$

$$\approx \frac{1}{K}\left[ s\left(\frac{x-t}{\rho}\right) s\left(\frac{\pi}{\sqrt{3}} \frac{x-\mu}{\sqrt{\sigma}}\right) \right]_{-\infty}^{z} - \frac{1}{K} \int_{-\infty}^{z} N\left(x\Big|t, \frac{\pi^2}{3}\rho^2\right) N(x|\mu,\sigma) dx$$

where the third line makes use of integration by parts, and the fourth line relies upon the logistic-matching trick (twice).

The first term in the right-hand side of Equation 37 is simply given by:

$$\frac{1}{K}\left[ s\left(\frac{x-t}{\rho}\right) s\left(\frac{\pi}{\sqrt{3}} \frac{x-\mu}{\sqrt{\sigma}}\right) \right]_{-\infty}^{z} = \frac{1}{K} s\left(\frac{z-t}{\rho}\right) s\left(\frac{\pi}{\sqrt{3}} \frac{z-\mu}{\sqrt{\sigma}}\right) \qquad (38)$$

Evaluating the second term derives from noting that the product of two gaussian densities is also a gaussian density. Let $f(x)$ be the logarithm of the product of two gaussian densities with means $\mu_1$ and $\mu_2$ and variances $\sigma_1$ and $\sigma_2$:

$$f(x) = \ln N(x|\mu_1,\sigma_1) + \ln N(x|\mu_2,\sigma_2)$$

$$= -\ln 2\pi - \frac{1}{2}\ln(\sigma_1 \sigma_2) - \frac{1}{2\sigma_1}(x-\mu_1)^2 - \frac{1}{2\sigma_2}(x-\mu_2)^2$$

$$= \ln N(x|\mu_{12},\sigma_{12}) - \frac{1}{2}\ln 2\pi - \frac{1}{2}\ln(\sigma_1 + \sigma_2) - \frac{1}{2}\frac{(\mu_1+\mu_2)^2}{\sigma_2+\sigma_1} \qquad (39)$$

$$\text{with} \begin{cases} \mu_{12} = \sigma_{12}\left(\sigma_1^{-1}\mu_1 + \sigma_2^{-1}\mu_2\right) \\ \sigma_{12} = \dfrac{1}{\sigma_1^{-1} + \sigma_2^{-1}} \end{cases}$$

Thus the second term in the RHS of Eq. 37 writes:

$$\frac{1}{K}\int_{-\infty}^{z} N\left(x\bigg|t,\frac{\pi^2}{3}\rho^2\right)N(x|\mu,\sigma)dx = \frac{\omega}{K}\int_{-\infty}^{z} N(x|\eta,\upsilon)dx$$

$$= \frac{\omega}{K} s\left(\frac{\pi}{\sqrt{3}}\frac{z-\eta}{\sqrt{\upsilon}}\right)$$

$$\text{with } \begin{cases} \omega = \dfrac{1}{2\pi}\dfrac{1}{\sqrt{\dfrac{\pi^2\rho^2}{3}+\sigma}}\exp-\dfrac{1}{2}\dfrac{(t+\mu)^2}{\dfrac{\pi^2\rho^2}{3}+\sigma} \\ \eta = \upsilon\left(\dfrac{3t}{\pi^2\rho^2}+\dfrac{\mu}{\sigma}\right) \\ \upsilon = \dfrac{1}{\dfrac{3}{\pi^2\rho^2}+\sigma^{-1}} \end{cases} \quad (40)$$

Inserting Equations 38 and 40 into Equation 37 yields the cumulative density function of the skewed normal distribution

$$P_s(z) = \frac{1}{K}s\left(\frac{z-t}{\rho}\right)s\left(\frac{\pi}{\sqrt{3}}\frac{z-\mu}{\sqrt{\sigma}}\right) - \frac{\omega}{K}s\left(\frac{\pi}{\sqrt{3}}\frac{z-\eta}{\sqrt{\upsilon}}\right)$$

$$K \approx s\left(\frac{\mu-t}{\sqrt{\rho^2+\dfrac{3}{\pi^2}\sigma}}\right) \quad (41)$$

where the expression for the normalization constant $K$ follows from the analytical approximation to the expected sigmoid (cf. Equation 8 above).

From Equation 41, one can check that $P_s(z)\xrightarrow{t\to-\infty} s\left(\dfrac{\pi}{\sqrt{3}}\dfrac{z-\mu}{\sqrt{\sigma}}\right)$, essentially because $s\left(\dfrac{z-t}{\rho}\right)\xrightarrow{t\to-\infty}1$ and $\omega\xrightarrow{t\to-\infty}0$. This is important, because this means that the skewed-normal distribution tends to the gaussian distribution $N(x|\mu,\sigma)$ when $t\to-\infty$, i.e. when the sigmoid reaches its plateau far away on the left of the mode of the gaussian density.

## 3.2 The expected log-sum of Bernouilli variables

Let $\{b_i\}_{i=1,\ldots,N}$ be a collection of $N$ independent Bernouilli random variables with sufficient statistics $\lambda_i = p(b_i = 1) = E[b_i]$. We are looking for an approximation of the following quantity: $\langle \log(\sum_i b_i + 1) \rangle$. Although apparently arbitrary, our interest here is justified by the fact that this quantity naturally arises in the context of (variational) Bayesian inference on Dirichlet processes (Blei and Jordan, 2005), which are quite popular in the context of non-parametric data clustering problems.

First, note the two first moments of the sum of Bernouilli variables are given by:

$$\begin{cases} E\left[\sum_i b_i\right] = \sum_i \lambda_i \\ V\left[\sum_i b_i\right] = \sum_i V[b_i] = \sum_i \lambda_i (1 - \lambda_i) \end{cases} \qquad (42)$$

Since the $b_i$'s are binary variables, their sum is positive. This means that there exists a variable $x$ such that: $\sum_{i=1}^{N} b_i = \exp(x)$. Furthermore, when $N$ gets large ($N \gg 1$), $\sum_i b_i$ converges to a continuous variable. Thus, the expected log sum can be approximated as follows:

$$\left\langle \log\left(\sum_i b_i + 1\right) \right\rangle \approx \left\langle \log(\exp(x) + 1) \right\rangle = -\left\langle \log s(-x) \right\rangle \qquad (43)$$

where the expectation on the right hand side of Equation 43 is taken under the (yet unknown) probability density function of $x$. Assuming $x$ follows a Gaussian distribution with mean $\mu$ and variance $\Sigma$, $\exp(x)$ will follow a log-normal distribution, whose first two moments have to match those of $\sum_i b_i$. This yields:

$$\begin{cases} \mu = 2\log \sum_i \lambda_i - \frac{1}{2}\log\left(\sum_i \lambda_i(1-\lambda_i) + \left(\sum_i \lambda_i\right)^2\right) \\ \Sigma = \log\left(\sum_i \lambda_i(1-\lambda_i) + \left(\sum_i \lambda_i\right)^2\right) - 2\log \sum_i \lambda_i \end{cases} \quad (44)$$

A fixed form approximation to the expected log sum of Bernouilli variables can then be derived from the application of equations 12 and 15:

$$\left\langle \log\left(\sum_i b_i + 1\right) \right\rangle \approx -\langle \log s(-x) \rangle$$
$$\approx \log\left(1 + \exp\left(\frac{\mu - b\Sigma^c}{\sqrt{1+a\Sigma^d}}\right)\right) \quad (45)$$

where $\mu$ and $\Sigma$ are given by equation 44.

### 3.3 Moments of absolute values of a Gaussian variable

First, let us derive an approximation to the absolute-value mapping $g(x) \equiv |x|$ as a limit of a continuous (and differentiable) function. First, note that the derivative of $g$ is the so-called "step function", which can be seen as the limit of the following translated sigmoid mapping $s(x/\rho)$:

$$\frac{\partial}{\partial x} g(x) = 2 \lim_{\rho \to 0} s\left(\frac{x}{\rho}\right) - 1 \quad (46)$$

This is interesting, because the absolute-value can now be approximated as the limit of the antiderivative $g_\rho(x)$ of the (translated) sigmoid mapping:

$$g(x) = \lim_{\rho \to 0} g_\rho(x)$$
$$g_\rho(x) = x - 2\rho \log s\left(\frac{x}{\rho}\right) \quad (47)$$
$$\frac{\partial}{\partial x} g_\rho(x) = 2s\left(\frac{x}{\rho}\right) - 1$$

The accuracy of the approximation ($x \to g_\rho(x)$) to the absolute-value mapping ($x \to |x|$) is depicted on Figure 8 below:

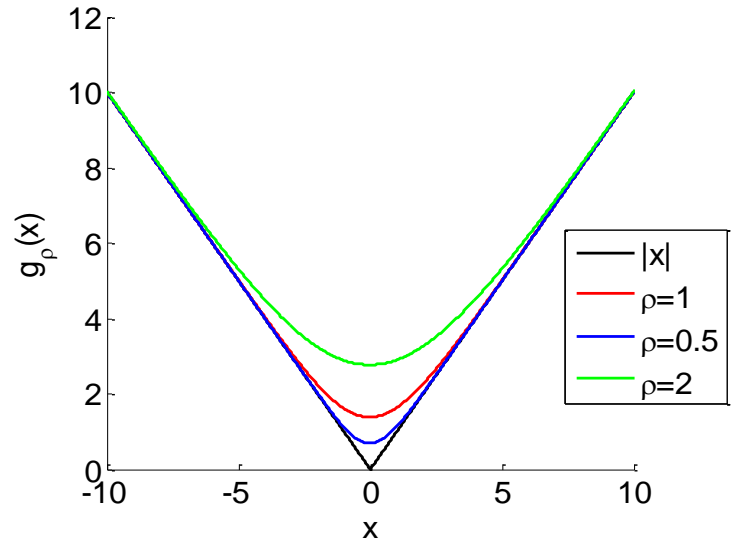

Figure 8: Fixed-form approximation to $|x|$.

In turn, one can derive an approximation to the expected absolute value of a Gaussian variable as follows:

$$\begin{aligned} E[|x|] &= E[g(x)] \\ &\approx \lim_{\rho \to 0} E[g_\rho(x)] \\ &= E[x] - 2 \lim_{\rho \to 0} \rho \, E\left[\log s\left(\frac{x}{\rho}\right)\right] \\ &= \mu - 2 \lim_{\rho \to 0} \rho \log s\left(\frac{\rho^{-1}\mu + b\rho^{-2c}\Sigma^c}{\sqrt{1 + a\rho^{-2d}\Sigma^d}}\right) \end{aligned} \qquad (48)$$

where are set according to the above numerical simulations on the expected log-sigmoid mapping (cf. equation 12).

Equation 48 can be used with an appropriately small $\rho$ to provide a semi-analytical approximation to $E[|x|]$.

## 4. Discussion

In conclusion, we have provided semi-analytical approximations of moments of Gaussian random variables passed through sigmoid or softmax mappings. These approximations are accurate (they yield 5% error at most) and verify basic requirements of sigmoid mappings, including boundedness (effective sigmoids should be positive and smaller than one), effect of $\mu$ on the effective sigmoid's inflexion point, effect of $\Sigma$ on the effective sigmoid's slope, etc...

We also have highlighted a few niche applications of these approximations, which arise in the context of, e.g., drift-diffusion models of decision making or non-parametric data clustering approaches. Note that these were only provided as examples of how to use these approximations as efficient alternatives to more tedious derivations that would be needed if one was to approach the underlying mathematical issues in a more formal way.

Note that, in the context of our own work, approximations of this sort have already found their way in many probabilistic models of learning and decision making (Daunizeau et al., 2010; Devaine et al., 2014; Lebreton et al., 2015; Mathys et al., 2011). We hope that this technical note will be helpful to modellers facing similar mathematical issues, although maybe stemming from different academic prospects.